\pdfoutput=1

\documentclass[11pt]{article}

\usepackage[]{EMNLP2023}

\usepackage{times}
\usepackage{latexsym}
\usepackage{amsmath}
\usepackage[T1]{fontenc}

\usepackage[utf8]{inputenc}

\usepackage{microtype}
\usepackage{CJKutf8}
\usepackage{inconsolata}
\usepackage{amsmath, amssymb}
\usepackage{makecell}
\usepackage{multirow}
\usepackage{graphicx}
\usepackage{subfigure}
\usepackage{changes}
\title{Improving speech translation by fusing speech and text}

\author{Wenbiao Yin\textsuperscript{1}, Zhicheng Liu\textsuperscript{2}, Chengqi Zhao\textsuperscript{2}, Tao Wang\textsuperscript{2}, Jian Tong\textsuperscript{2}, Rong Ye\textsuperscript{2} \\
    \textsuperscript{1}Nanjing University  \textsuperscript{2}ByteDance \\
    wenbiaoyin@smail.nju.edu.cn\\
    \{liuzhicheng.lzc, zhaochengqi.d, wangtao.960826, tongjian, yerong\}@bytedance.com }

\begin{document}
\maketitle
\begin{abstract}
In speech translation, leveraging multimodal data to improve model performance and address limitations of individual modalities has shown significant effectiveness. In this paper, we harness the complementary strengths of speech and text, which are disparate modalities. We observe three levels of modality gap between them, denoted by Modal input representation, Modal semantic, and Modal hidden states. To tackle these gaps, we propose \textbf{F}use-\textbf{S}peech-\textbf{T}ext (\textbf{FST}), a cross-modal model which supports three distinct input modalities for translation: speech, text, and fused speech-text. We leverage multiple techniques for cross-modal alignment and conduct a comprehensive analysis to assess its impact on speech translation, machine translation, and fused speech-text translation. We evaluate FST on MuST-C, GigaST, and newstest benchmark. Experiments show that the proposed FST achieves an average 34.0 BLEU on MuST-C En$\rightarrow$De/Es/Fr (vs SOTA +1.1 BLEU). Further experiments demonstrate that FST does not degrade on MT task, as observed in prior works. Instead, it yields an average improvement of 3.2 BLEU over the pre-trained MT model.

\end{abstract}

\section{Introduction} \label{Introduction}
Speech translation (ST) accepts speech signals as the input and outputs target translation. The field of speech translation can be broadly categorized into cascade system and end-to-end speech translation (E2E ST). Cascade system\cite{sperber-etal-2017-neural,zhang-etal-2019-lattice,9413719} usually combines automatic speech recognition (ASR) and machine translation (MT), the MT subsystem uses ASR transcripts as input, which provide clear expression but may contain errors stemming from ASR. While E2E ST\cite{tang-etal-2021-improving,fang-etal-2022-stemm,ye-etal-2022-cross} can directly map speech signals to the target translation, thus avoiding the problem of error propagation. 

\begin{figure}[t]
\centering
\includegraphics[width= 1.1\linewidth]{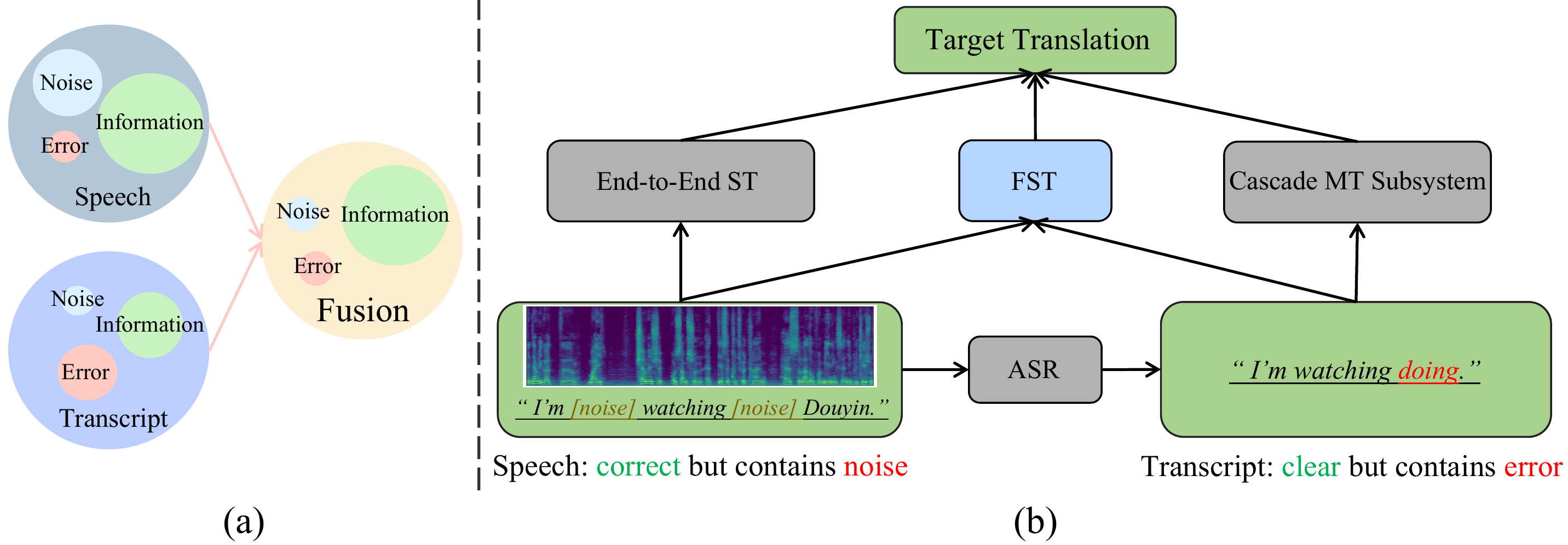}
\caption{\textbf{Left}: Data distribution of speech and transcript. \textbf{Right}: The pipeline of our proposed FST.}
\label{Distribution}
\end{figure}

We observe a noticeable modality distribution difference between speech and ASR transcript as shown in Figure \ref{Distribution}a. Compared to ASR transcript, speech signals contain richer information, such as paralinguistics and speaker characteristics; they are harder to model and more susceptible to noise. In summary, speech signals contain accurate and abundant information and substantial noise, while ASR transcripts are clear but prone to errors. 

As shown in Figure \ref{Distribution}b, the speech signal conveys the message "I'm watching Douyin.", but the speech signal contains numerous blank segments and background noise. E2E ST may encounter challenges in accurately extracting information directly from speech, particularly in the presence of noise, without compromising the quality of the translation. Meanwhile, the ASR model mistakenly transcribes the speech as "I'm watch doing." The MT subsystem may identify the error in the word "doing" as the decoder assigns low probabilities to candidate words. However, the MT subsystem lacks additional information to correct the error.  

Inspired by these findings, we propose a model that fuses speech and ASR transcript as input to leverage their complementary strengths. As shown in Figure \ref{Distribution}b, our model supports three different modalities of input for translation: speech, text, and fused speech-text. Our model leverages corresponding speech representation to rectify errors when our model identifies the error in the word "doing" (as the decoder assigns low probabilities to candidate words). Furthermore, we can enhance the training of our model by utilizing additional MT data.

However, speech and text are disparate modalities, we observe three levels of the modality gap between them, which impede the integration of speech and text: \textbf{1. Modal input representation}: The length of speech input representation obtained from speech pre-trained model frequently exceeds that of the corresponding transcript input representation by an order of magnitude or more. Furthermore, it is worth noting that while speech input representation is in continuous space, text input representation is in discrete space. \textbf{2. Modal semantic}: When using speech and golden transcript as input, the model relies heavily on the golden transcript information and neglects the speech information. This behavior is because the golden transcript is more straightforward and contains precise semantic information, adequate for achieving high translation quality. In contrast, speech representations obtained through the pre-trained model include four types of information\cite{chen2022wavlm}: content, semantics, paralinguistics, and speaker characteristics, which is harder to model. However, when using speech and ASR transcript as input, solely relying on textual information is insufficient to achieve good translation quality. We expect the model to incorporate more speech information. \textbf{3. Modal hidden states}: It is observed that diverse modality inputs lead to distinctive hidden states in the encoder and distinct distributions in the decoder.

We propose several methods to bridge the modality gap to integrate speech and text better. \textbf{1.} We utilize a convolutional layer to shrink the length of speech representations by a factor of $\mu$, which enable speech and text to have a comparable scale. Additionally, we explore mapping continuous speech representation to a discrete space using a codebook to align with text representation. \textbf{2.} We utilize an open-source ASR model to construct a dataset of <speech, ASR transcript> pair and employe prompt tags (<golden>/<asr>) to indicate whether the transcript contains errors. The prompt tags implicitly guide the model to utilize more speech information when the transcript is inaccurate. Meanwhile, we utilize the contrastive learning method to reduce the gap between the model semantic of speech and its corresponding transcript. \textbf{3.} We utilize Cross-Attentive Regularization (CAR) to align the states of the encoder and Jensen-Shannon Divergence (JSD) loss to align the distribution of the decoder.

Our contributions are summarized as follows:
\begin{itemize}
\item[$\bullet$]We propose a model that fuses speech and text to improve speech translation, which supports three different modalities of input for translation: speech, text, and fused speech-text.
\end{itemize}
\begin{itemize}
\item[$\bullet$]To fuse speech and text as input and leverage their complementary strengths, we conducted a comprehensive analysis of the modality gap between speech and text. We propose targeted improvements to bridge the modality gap between speech and text.
\end{itemize}
\begin{itemize}
\item[$\bullet$]
Our experiments show that our model achieves an average 34.0 BLEU on MuST-C En$\rightarrow$De/Es/Fr (vs SOTA +1.1 BLEU); FST achieves an average improvement of 3.2 BLEU over the pre-trained MT model on MuST-C.
\end{itemize}

\section{Related Work}
\noindent \textbf{Cascade ST}
Cascade ST, achieved by concatenating ASR and MT components, has been extensively employed in commercial speech translation systems. However, cascade ST is vulnerable to challenges such as error propagation and high latency. To overcome the error propagation, (\citealp{1566492,beck-etal-2019-neural,sperber-etal-2019-self}) proposed to feed the MT system with ASR data structures; (\citealp{peitz-etal-2012-spoken,cheng-etal-2019-breaking,di-gangi-etal-2019-robust}) proposed to make MT robust to ASR errors, for instance by training it on parallel data incorporating factual or emulated ASR errors.

\noindent \textbf{End-to-end ST} 
To overcome the error propagation and high latency in the cascade ST systems, \cite{berard:hal-01408086,duong-etal-2016-attentional} proposed an end-to-end architecture for speech translation, which has attracted extensive attention (\citealp{Vila2018EndtoEndST,salesky-etal-2019-fluent,gangi19_interspeech,inaguma-etal-2021-source,zhao-etal-2021-mutual}). However, it is difficult to train an end-to-end speech translation model directly, primarily due to the inherent variability and complexity of speech signals and the scarcity of high-quality speech-translation datasets. Some training methods like pretraining (\citealp{Weiss2017SequencetoSequenceMC,10.1109/ICASSP.2018.8461690,bansal-etal-2019-pre,wang2020bridging,9415058}), multi-task learning(\citealp{le-etal-2020-dual,9414159,ye2021end,tang-etal-2022-unified}), data augmentation (\citealp{Park2019SpecAugmentAS,jia2019leveraging,bahar-etal-2019-using,pino20_interspeech}), meta-learning (\citealp{indurthi2019data}), contrastive learning (\citealp{li-etal-2021-unimo,ye-etal-2022-cross}), knowledge distillation (\citealp{liu19d_interspeech,tang-etal-2021-improving}) and curriculum learning (\citealp{Kano2017StructuredBasedCL,wang-etal-2020-curriculum}), are proved to be effective.

\section{Methods}

\subsection{Problem Formulation}
The speech translation corpus is usually comprised of triples that include speech, transcript, and target translation, which can be denoted as $D = {(s, x, y)}$. Here, $s$ is an audio sequence, $x$ is the corresponding transcript, and $y$ is the corresponding target translation.

\begin{figure*}[t]
\centering
\includegraphics[width=0.9 \linewidth]{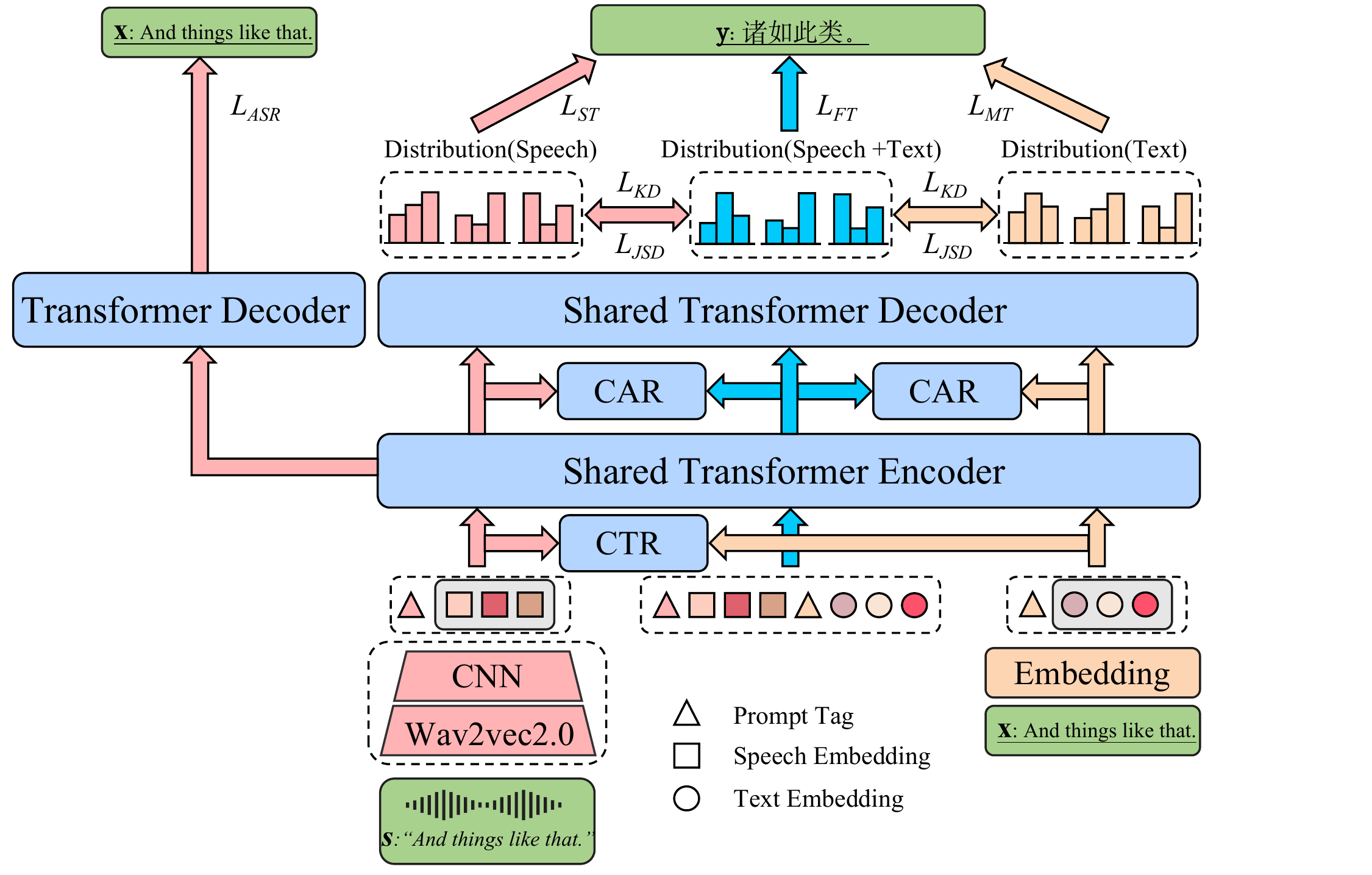}
\caption{Overview of our proposed FST. FT denotes fused speech-text input for translation; here, we use the prompt tag to fuse speech and text. CTR denotes cross-modal contrastive learning; CAR denotes cross-attentive regularization.}
\label{model}
\end{figure*}

\subsection{Model Framework}

As shown in Figure~\ref{model}, our model consists of four sub-modules: Speech Encoder, Speech-Text Fusion Module, Transformer Encoder, and Transformer Decoder. Our model supports three different modalities as input: speech, text, and fused speech-text, and to output target translation. We abbreviate these three different input modalities for translation as ST, MT, and FT(fused speech-text translation), respectively. 

\noindent \textbf{Speech Encoder} The speech encoder($S\mbox{-}Enc$) consists of Wav2vec2.0\cite{baevski2020wav2vec} and
two additional convolutional layers. The input is a raw waveform signal sampled at 16kHz, and Wav2vec2.0 is used to extract low-level speech representations from it. However, the speech representation is much longer than the text, exceeding it by over ten times. To reduce the difference in representation length between speech and text as much as possible, we use two additional convolutional layers with stride 2 to shrink the length by a factor of 4. A greater degree of downsampling would have led to information loss, while a lesser degree of downsampling could have resulted in modal misalignment and compromised performance. Denote $a=S\mbox{-}Enc(s)$ as the speech representation.

\vspace{12 pt}
To reduce the number of parameters and facilitate knowledge transfer, we share the \textit{Transformer encoder} and \textit{Transformer decoder} for ST, MT, and FT.

\noindent \textbf{Transformer Encoder and Transformer Decoder} The \textit{Transformer encoder} and \textit{Transformer decoder} are composed of $N_e$ transformer encoder layers and $N_d$ transformer decoder layers, respectively, with the same configuration as the original implementation\cite{vaswani2017attention}. For the MT task, the input of the \textit{Transformer encoder} is the embedding of transcript $e = Emb(x)$. For the ST task, the input is the audio output representation of the speech encoder $a$. For the FT task, the input is the fused speech-text representation $f$(see details in Section \ref{fused-speech-text-section}). The \textit{Transformer encoder} further extracts the high-level semantic hidden representations and facilitates knowledge sharing across the three modalities. The \textit{Transformer decoder} generates corresponding target translation for ST, MT, and FT. Besides, we use ASR task to improve translation performance. The training losses of ST, MT, FT, and ASR are as follows:
\begin{equation}
\begin{aligned}
\mathcal{L}_{ST} &= -\sum_{n} \log P(y_n |s_n) \\
\mathcal{L}_{MT} &= -\sum_{n} \log P(y_n |x_n) \\
\mathcal{L}_{FT} &= -\sum_{n} \log P(y_n |x_n,s_n) \\
\mathcal{L}_{ASR} &= -\sum_{n} \log P(x_n |s_n) \\
\end{aligned}
\end{equation}
$\mathcal{L}_{ST}$, $\mathcal{L}_{MT}$, $\mathcal{L}_{FT}$ and $\mathcal{L}_{ASR}$ are cross-entropy losses on <speech, target>, <transcript, target>, <speech, transcript, target> and <speech, transcript>, respectively.

\subsection{Fusing speech and text}\label{fused-speech-text-section}
As we mentioned in Section~\ref{Introduction}, to leverage the complementary strengths between speech and text, we propose the \textbf{F}use-\textbf{S}peech-\textbf{T}ext(\textbf{FST}) method. We first introduce FST in this section and later show how to bridge the modality gap between speech and text.

Here, we utilize an open-source ASR model to construct a <speech, ASR transcript, target> pair dataset. Given a speech-transcript-target pair $(s,x,y)$, the transcript $x$ could be an ASR transcript or golden transcript, and the fused speech-text representation $f$ is defined as:
\begin{equation}
\begin{aligned}
f = Concat(P^s,S\mbox{-}Enc(s),P^t,P^g,Emb(x))
\end{aligned}
\end{equation}
where $P^s$ and $P^t$ are prompt tags to identify whether the input modal is speech or text, and $P^g$ is a prompt tag (<golden>/<asr>) to indicate whether the transcript contains errors. The prompt tags implicitly guide the model to utilize more speech information when the transcript is inaccurate.


\subsection{Align speech and text with FST}

As we analyzed in Section~\ref{Introduction}, we observe three levels of the modality gap between speech and text: Modal input representation, Modal semantic, and Modal hidden states. We propose several methods to bridge the modality gap between speech and text. In addition to the speech subsampling and implicit guidance mentioned above, we introduce three additional methods in this section: Cross-modal Contrastive Learning, Cross-Attentive Regularization, Knowledge Distillation and Jensen-Shannon Divergence.

\noindent \textbf{Cross-modal Contrastive Learning}
Given a positive example of speech-transcript $(s,x)$ pair, we can get speech representation $a=S\mbox{-}Enc(s)$, transcript representation $e = Emb(x)$. We randomly pick a set of $B-1$ transcripts $\left \{ e_i^- \right \}_{i=1}^{B-1} $ from the same batch as negative examples. For speech representation $a$ and transcript representation $e$, we first average them in terms of the time dimension and apply the multi-class N-pair contrastive loss\cite{sohn2016improved}:
\begin{equation}
\begin{aligned}
u &= MeanPool(a) \\
v &= MeanPool(e) \\
\mathcal{L}_{CTR} = -\sum_{s,x}  \log &\frac{exp(sim(u,v)/\tau )}{\sum_{e_j \in \mathcal{A}} exp(sim(u,v(e_j))/\tau ) } 
\end{aligned}
\end{equation}
where $\mathcal{A}=\left \{ e \right \} \cup \left \{ e_i^- \right \}_{i=1}^{B-1}$, $\tau$ is the temperature hyper-parameter, and $sim$ is the cosine similarity function.

\noindent \textbf{Cross-Attentive Regularization} The cross-attentive regularization\cite{tang-etal-2021-improving} (CAR) can increase the similarity among distinct modalities. The essence of CAR is to use a similarity matrix to project a tensor sequence onto a space of equivalent length as another tensor sequence, then compute $L2$ loss between the two sequences. Here, we utilize the CAR to compute losses separately between speech representation $a$ and fused speech-text representation $f$ and between text representation $e$ and fused speech-text representation $f$. The CAR loss is defined as:

\begin{equation}
\begin{aligned}
    \mathcal{L}_{CAR} = \mathcal{L}_{CAR}(a,f) + \mathcal{L}_{CAR}(e,f) 
\end{aligned}
\end{equation}

\noindent \textbf{Knowledge Distillation and Jensen-Shannon Divergence}
The ST task is more difficult than the MT task since the speech signals are harder to model and more susceptible to noise. Previous works\cite{liu19d_interspeech,gaido-etal-2020-end,tang-etal-2021-improving} utilize knowledge distillation to facilitate knowledge acquisition by an ST model from a well-trained MT model. However, in our work, the transcript may contain errors; the accuracy of the FT task is usually much higher than the corresponding ST and MT task. We designate the FT task as the teacher while ST and MT tasks as the student, minimizing the loss between the student and teacher outputs. The KD loss is defined as:
\begin{equation}
\begin{aligned}
    \mathcal{L}_{KD}(a,f) &= - \sum_{n} q(y_n|f_n) \log p(y_n|a_n) \\
    \mathcal{L}_{KD}(e,f) &= - \sum_{n} q(y_n|f_n) \log p(y_n|e_n) \\
    \mathcal{L}_{KD} & = \mathcal{L}_{KD}(a,f) + \mathcal{L}_{KD}(e,f)
\end{aligned}
\end{equation}

Furthermore, we endeavor to optimize the congruity of the ultimate distributions of the three modalities through the minimization of the Jensen-Shannon Divergence\cite{lin1991divergence} (JSD) between the three output distributions, which is

\begin{equation}
\begin{aligned}
\mathcal{L}_{JSD} = \sum_{n} (&JSD \left \{ p(y_n | a_n) ||  p(y_n | ,f_n)\right \} \\
+ &JSD \left \{ p(y_n | e_n) ||  p(y_n | f_n)\right \}) 
\end{aligned}
\end{equation}

The final training objective is as follows:
\begin{equation}
\begin{aligned}
    \mathcal{L} =& \alpha \mathcal{L}_{ST} + \alpha \mathcal{L}_{MT} + (1-\alpha) \mathcal{L}_{KD} +\mathcal{L}_{FT}  \\
            +& \mathcal{L}_{ASR} + \beta \mathcal{L}_{CTR} + \gamma \mathcal{L}_{CAR} + \delta  \mathcal{L}_{JSD} 
\end{aligned}
\label{final_loss}
\end{equation} 
where $\alpha$, $\beta$, $\gamma$ and $\delta$ are predefined hyper-parameters.

\section{Experiments}

\subsection{Experimental setups}
\textbf{ST Datasets} We conduct experiments on MuST-C\footnote{We use v1.0 \url{https://ict.fbk.eu/must-c/}}\cite{di-gangi-etal-2019-must} and GigaST\footnote{\url{https://st-benchmark.github.io/resources/GigaST}}\cite{gigast}. MuST-C is a multilingual speech translation dataset that contains translations from English to 8 languages. MuST-C contains several hundred hours of audio recordings from English TED Talks; we conduct experiments on MuST-C En$\rightarrow$De/Es/Fr. We use the dev set for development and analysis and tst-COMMON set for test. GigaST is a large-scale pseudo speech translation dataset created by translating the text in GigaSpeech. We conduct experiments on GigaST En$\rightarrow$Zh, and test on in-house cgtn, zhiyuan and aiconf datasets. We utilize an open-source ASR model (whisper base.en
 \footnote{\url{https://huggingface.co/openai/whisper-base.en}}) to construct a dataset of <speech, ASR transcript> pair, the Word Error Rate (WER) of our constructed datasets are shown in the last column of Table~\ref{datasets}.

\begin{table}[t]
\resizebox{\linewidth}{!}{ 
\begin{tabular}{ccccccc}
\hline
& \multicolumn{2}{c}{ST} & \multicolumn{2}{c}{MT Train} & \multicolumn{1}{c}{MT Test} & \multirow{2}{*}{WER} \\
 & \#hours & \#sents & name & \#sents  & name & ~\\ \hline
\multicolumn{6}{l}{MuST-C} \\ \hline
En$\rightarrow$De & 408 & 234K & WMT17 & 4.5M & newstest2014 & 22.65\\
En$\rightarrow$Es & 504 & 270K & WMT17 & 4.1M & newstest2013 & 22.60\\
En$\rightarrow$Fr & 492 & 280K & WMT17 & 5.5M & newstest2014 & 22.08\\ \hline
\multicolumn{6}{l}{GigaST} \\ \hline
En$\rightarrow$Zh & 9,780 & 7,650K & in-house & 105M &  newstest2019/2020 & 21.98\\
\hline
\end{tabular}}
\caption{Statistics of all datasets.}
\label{datasets}
\end{table}

\noindent \textbf{MT Datasets} Our model allows us to use the external MT dataset for further training. We introduce external WMT datasets for En$\rightarrow$De/Es/Fr and in-house MT dataset for En$\rightarrow$Zh. The detailed statistics of all datasets are shown in Table~\ref{datasets}.

\begin{table*}[t]
\centering
\begin{tabular}{l|cccc|ccc|c}
\hline
\multirow{2}{*}{Models} & \multicolumn{4}{c}{External Data} \vline & \multicolumn{4}{c}{BLEU} \\ 
~ & Speech & Text & ASR & MT & De & Es & Fr & Avg. \\ \hline
\multicolumn{9}{l}{Cascade Model} \\ \hline
Espnet\cite{inaguma-etal-2021-source} &  - & - & - & - & 23.6 & - & 33.8 & - \\
W2V2-Transformer\cite{fang-etal-2022-stemm}  & $\checkmark$  & - & - & $\checkmark$ & 26.9 & 30.0 & 36.6 & 31.2 \\
\cite{ye2021end} &  - & - & $\checkmark$ & $\checkmark$ & 25.2 & - & 34.9 & - \\
\cite{xu-etal-2021-stacked} &  - & - & $\checkmark$ & $\checkmark$ & 28.1 & - & - & - \\
\hline
\multicolumn{9}{l}{End-to-End Model} \\ \hline
MTL\cite{9415058} & - & - & - & $\checkmark$ & 23.9 & 28.6 & 33.1 & 28.5 \\
FAT-ST\cite{zheng2021fused} & $\checkmark$ & $\checkmark$ & $\checkmark$ & $\checkmark$ & 25.5 & 30.8 & - &  - \\
JT-S-MT\cite{tang-etal-2021-improving} &  - & - & - & $\checkmark$ & 26.8 & 31.0 & 37.4 & 31.7 \\
Chimera\cite{han2021learning} & $\checkmark$ & - & - & $\checkmark$  & $\text{27.1}^{\dagger}$  & 30.6 & 35.6 & 31.1 \\
XSTNet\cite{ye2021end} & $\checkmark$ & - & - & $\checkmark$ & 27.1 & 30.8 & 38.0 & 32.0 \\
SATE\cite{xu-etal-2021-stacked} & - & - & $\checkmark$ & $\checkmark$ & $\text{28.1}^{\dagger}$ & - & - & - \\
STEMM\cite{fang-etal-2022-stemm} & $\checkmark$ & - & - & $\checkmark$ & 28.7 &  31.0 & 37.4 & 32.4 \\
TaskAware\cite{9414703}&  - & - & $\checkmark$ & $\checkmark$ & 28.9 & - & - & - \\
STPT\cite{tang-etal-2022-unified} & $\checkmark$ & $\checkmark$ & $\checkmark$ & $\checkmark$ & - & 33.1 & \textbf{39.7} & - \\
ConST\cite{ye-etal-2022-cross} &  $\checkmark$ & - & - & $\checkmark$ & 28.3 & 32.0 & 38.3 & 32.9 \\ \hline
\textbf{FST-ST} & - & - & - & $\checkmark$ & 27.7 & 32.4 & 37.2 & 32.4 \\
\textbf{FST-FT} &  - & - & - & $\checkmark$ & \textbf{29.2} & \textbf{33.9} & 38.9 & \textbf{34.0} \\ \hline

\end{tabular}
\caption{Case-sensitive detokenized BLEU scores on MuST-C tst-COMMON set. "Speech" denotes unlabeled audio data, "Text" denotes unlabeled text data, \textit{e.g.} Europarl V7\cite{koehn2005europarl}, CC25\cite{liu2020multilingual}, $\dagger$ use 40M OpenSubtitles\cite{lison-tiedemann-2016-opensubtitles2016} as external MT data. }
\label{mustc_result}
\end{table*}

\begin{table*}[t]
\centering
\resizebox{\linewidth}{!}{ 
\begin{tabular}{c|ccc|ccc|ccc|c}
\hline
\multirow{2}{*}{Models} & \multicolumn{3}{c}{De}  \vline & \multicolumn{3}{c}{Es} \vline & \multicolumn{3}{c}{Fr}  \vline & \multirow{2}{*}{Avg.}\\
~ & newstest & asr & golden & newstest & asr & golden & newstest & asr & golden & ~\\ \hline
Pre-trained 6E6D $\text{MT}^\sharp$ & 23.7 & 24.3 & 30.4 & 31.8 & 29.5 & 34.2 & 34.7 & 32.2 & 39.9 & 31.2\\ 
Pre-trained 24E6D $\text{MT}^\sharp$ & \textbf{27.5} & 25.7 & 31.9 & \textbf{34.8} & 32.0 & 37.0 & \textbf{38.1} & 34.2 & 41.5 & 33.6\\
STEMM \cite{fang-etal-2022-stemm} & - & - & 31.5 & - & - & - & - & - & - & -\\
\textbf{FST-MT} & 25.0 & \textbf{28.6} & \textbf{34.2} & 32.9 & \textbf{33.5} & \textbf{37.7} & 35.9 & \textbf{37.7} & \textbf{44.1} & \textbf{34.4}\\ \hline
\end{tabular}}
\caption{Case-sensitive detokenized BLEU scores on MuST-C tst-COMMON set and newstest. "asr" denotes use the ASR transcript as input, "golden" denotes use the golden transcript as input. $\sharp$ are trained with the same external MT data and MuST-C <transcript, target> pair data.}
\label{mt_result}
\end{table*}

\noindent\textbf{Model Configuration}
For the speech encoder, we use Wav2vec2.0\footnote{\url{https://dl.fbaipublicfiles.com/fairseq/wav2vec/wav2vec_small.pt}} following the
base configuration, which is only pre-trained on Librispeech\cite{7178964} without any finetuning. Two layers of CNNs after the Wav2vec2.0 with kernel size 5, stride size 2, padding 2, and hidden dimension 1024. The transformer encoder and decoder follow the base configuration, with hidden size $h^d = 512$, 8 attention heads, and 2048 FFN hidden states. We use $N_e=6$ transformer encoder layers and $N_d=6$ transformer decoder layers.

\noindent\textbf{Experiment Details}  We pre-trained the MT model on the external MT dataset; the learning rate is 5e-4. The FST model is initialized with a pre-trained MT model, and the learning rate is 6e-5. We use the raw 16kHZ speech as input and jointly tokenize the bilingual text using SentencePiece\cite{kudo-richardson-2018-sentencepiece}. We use an Adam optimizer with $\beta_1=0.9$, $\beta_2=0.98$, and 20k warm-up updates. The dropout is set to 0.15, and the value of label smoothing is set to 0.1. For the training loss, we set 
weight of $ \mathcal{L}_{ST}$ and $ \mathcal{L}_{MT}$ $\alpha=0.8$, contrastive temperature $\tau=0.02$ and 
weight of $ \mathcal{L}_{CTR}$ $\beta=1.0$, weight of $ \mathcal{L}_{CAR}$ $\gamma=0.02$ and weight of $ \mathcal{L}_{JSD}$ $\delta=1.0$. We use sacreBLEU\footnote{\url{https://github.com/mjpost/sacrebleu}, sacreBLEU signature: nrefs:1 | bs:1000 | seed:12345 | case:mixed | eff:no | tok:13a | smooth:exp | version:2.0.0}\cite{post-2018-call} to evaluate case-sensitive detokenized BLEU.

\begin{table*}[t]
\resizebox{\linewidth}{!}{ 
\begin{tabular}{l|cc|ccc|ccc|ccc}
\hline
\multirow{2}{*}{Models} & \multirow{2}{*}{newstest2019} & \multirow{2}{*}{newstest2020} & \multicolumn{3}{c}{cgtn}  \vline & \multicolumn{3}{c}{zhiyuan}  \vline & \multicolumn{3}{c}{aiconf} \\
~ & ~ & ~ & asr & speech & fused & asr & speech & fused & asr & speech & fused \\ \hline
Pre-trained 6E6D $\text{MT}^\sharp $ & 36.7 & 43.0 & 29.7 & - & - & 28.2 & - & - & 30.5 & - & - \\
Pre-trained 24E6D $\text{MT}^\sharp $ & \textbf{40.1} & \textbf{46.6} & \textbf{32.6} & - & - & \textbf{29.5} & - & - & \textbf{33.6} & - & - \\ \hline
SA-CTR \cite{li-etal-2022-vision}& 36.8 & 43.1 & 31.0 & 31.8 & 31.0 & 28.3 & 29.4 & 29.1 & 31.6 & 32.4 & 32.1 \\
Codebook-Gumbel\cite{Baevski_Schneider_Auli_2020}  & 36.8 & 43.0 & 30.9 & 30.0 & 31.7 & 28.1 & 28.4 & 28.9 & 31.7 & 31.6 & 32.5 \\
Codebook-K-means\cite{Baevski_Schneider_Auli_2020} & 36.9 & 43.2 & 31.0 & 31.3 & 31.8 & 28.2 & 29.3 & 29.0 & 31.5 & 31.7 & 32.3 \\ \hline
\textbf{FST} & 38.3 & 44.2 & 31.6 & \textbf{33.3} & \textbf{33.9} & 28.4 & \textbf{30.2} & \textbf{30.5} & 32.4 & \textbf{34.0} & \textbf{34.5} \\ \hline
\end{tabular}}
\caption{Case-sensitive detokenized BLEU scores on En$\rightarrow$Zh test sets. $\sharp$ are trained with the same MT data and GigaST <transcript, target> pair data. "asr" denotes use the ASR transcript, "fused" denotes use ASR transcript and speech as input. Hubert large is used to extract speech features on the GigaST En$\rightarrow$Zh.}
\label{en_zh_test}
\end{table*}

\subsection{Baseline systems}
We compare our method with several strong ST systems, including Espnet\cite{inaguma-etal-2021-source}, W2V2-Transformer\cite{fang-etal-2022-stemm}, MTL\cite{9415058}, FAT-ST\cite{zheng2021fused}, JT-S-MT\cite{tang-etal-2021-improving}, Chimera\cite{han2021learning}, XSTNet\cite{ye2021end}, SATE\cite{xu-etal-2021-stacked}, STEMM\cite{fang-etal-2022-stemm}, TaskAware\cite{9414703}, STPT\cite{tang-etal-2022-unified}, ConST\cite{ye-etal-2022-cross}.

Besides, we implement several methods for fusing speech and text modalities. The only difference between our approach and others is the specific method for fusing speech and text. SA-CTR: our implementation involved drawing inspiration from the image and text fusion techniques employed in \citeposs{li-etal-2022-vision} to propose a method for fusing speech and text. Codebook-Gumbel and Codebook-K-means: we utilize the method from \citet{Baevski_Schneider_Auli_2020} to map the speech representation $e$ onto a discrete space using a codebook; then we concatenate discrete speech representation and text representation as ours.

\subsection{Main Results}
\textbf{Comparison with End-to-End Baselines}
As shown in Table~\ref{mustc_result}, we compare our model with several strong end-to-end baselines. Many existing works rely on additional auxiliary data for better performance, \textit{e.g.} large-scale MT data and unlabeled audio data. In the table, we provide a summary of the auxiliary data employed by these baselines, with a $\checkmark$ denoting its usage in the corresponding column. Most previous strong baselines have leveraged two or more additional auxiliary datasets, such as STPT\cite{tang-etal-2022-unified} and ConST\cite{ye-etal-2022-cross}, whereas our model only utilizes additional MT data. Our E2E FST-ST achieves comparable results with the previous best models. When fusing speech and asr transcript as input, our FST-FT outperforms SOTA by an average of 1.1 BLEU on MuST-C.

\begin{table}[t]
\centering
\resizebox{\linewidth}{!}{ 
\begin{tabular}{l|cccc}
\hline
Task & MT & MT & ST & FT \\
Config. & newstest & asr & speech & fused \\ \hline

FST & 25.0 & 28.6 & 27.7 & 29.2 \\
\quad $-\mathcal{L}_{ASR}$ & 24.9 & 28.5 & 27.4 & 28.9 \\
\quad $-\mathcal{L}_{ASR}-\mathcal{L}_{KD} - \mathcal{L}_{JSD}$& 25.2 & 28.0 & 27.1 & 28.8 \\
\quad $-\mathcal{L}_{ASR}-\mathcal{L}_{KD} - \mathcal{L}_{JSD}-\mathcal{L}_{CTR} -\mathcal{L}_{CAR}$& 24.8 & 27.6 & 26.4 & 28.5 \\ \hline
\end{tabular}}
\caption{BLEU scores on MuST-C En$\rightarrow$De tst-COMMON set and newstest set by removing individual losses.}
\label{ablation_study}
\end{table}

\noindent \textbf{Comparison with Cascade Baselines}
We compare our model with several strong cascade systems. W2V2-Transformer, \citet{ye2021end} and \citet{xu-etal-2021-stacked} provided three strong cascade systems trained using MuST-C and external ASR and MT data. As shown in  Table~\ref{mustc_result}, our end-to-end FST-ST achieves comparable results with these strong cascade models, while our FST-FT significantly outperforms these strong cascade models.

\noindent \textbf{Comparison with MT Baselines}
Our model is initialized with a pre-trained MT model and then jointly trained on multiple tasks. Previous work has encountered catastrophic forgetting problems on MT task during joint training\cite{fang-etal-2022-stemm}, which significantly degrades performance on MT tasks. We evaluate our model on the MT task and show the result in Table~\ref{mt_result} and Table~\ref{en_zh_test}. Our model achieves significant improvement on the MT task instead of a decline in performance. Our model even outperforms pre-trained 24E6D (24 transformer encoder layers and 6 transformer decoder layers) MT on MuST-C. However, with the increase of training data, our model performs lower than pre-trained 24E6D MT when using ASR transcript as input (En$\rightarrow$Zh). Nevertheless, our model can still outperform pre-trained 24E6D MT when fusing speech and text as input on En$\rightarrow$Zh.

\subsection{Ablation Study}
As shown in Equation~\ref{final_loss}, our training objective contains eight terms. In addition to the cross-entropy objective $\mathcal{L}_{ST},\mathcal{L}_{MT},\mathcal{L}_{FT}$, we investigate the effects of the other auxiliary training objectives. By gradually removing each loss, Table~\ref{ablation_study} shows the improvements brought by each auxiliary training objective.

\begin{table}[t]
\centering
\resizebox{\linewidth}{!}{ 
\begin{tabular}{l|cccc}
\hline
Task & MT & MT & ST & FT \\
Model & newstest & asr & speech & fused \\ \hline
Wav2vec2.0 & 25.0 & 28.6 & 27.7 & 29.2 \\
HuBERT Large & 25.0 & 28.4 & 29.3 & 29.7 \\
HuBERT Extra Large & 25.1 & 28.4 & 30.0 & 30.3 \\ \hline
\end{tabular}
}
\caption{The impact of speech pre-trained model on MuST-C En$\rightarrow$De tst-COMMON set and newstest set.}
\label{speech_model}
\end{table}

\section{Analysis}

\subsection{Impact of Different Fusion Methods}
We compare our model with different fusion methods, such as SA-CTR and STEMM. SA-CTR proposed a selective attention and gated fusion mechanism to fuse two different modalities; STEMM proposed the speech-text manifold mixup to mix up the representation sequences of different modalities. Our model achieves better results than theirs by utilizing a prompt-based approach. As Section~\ref{Introduction} mentions, speech input representation is in continuous space, while text input representation is in discrete space. We utilize the method from \citet{Baevski_Schneider_Auli_2020} to map the speech representation onto a discrete space using a codebook. However, Codebook-Gumbel and Codebook-K-means led to a decrease in BLEU score (shown in Table~\ref{en_zh_test}). Our conjecture here is that the existing speech pre-trained models extract continuous features that undergo a mapping process to a discrete space, resulting in the loss of audio information and a subsequent reduction in the BLEU score. Nonetheless, if the speech pre-trained models can extract high-quality discrete features, it is plausible that such discrete features could enhance performance.

\subsection{Impact of Speech Pre-trained Model}
We further explore the impact of the speech pre-trained model. Here, we report the results of Wav2vec2.0, HuBERT Large\footnote{\url{https://dl.fbaipublicfiles.com/hubert/hubert_large_ll60k.pt}} and HuBERT Extra Large\footnote{\url{https://dl.fbaipublicfiles.com/hubert/hubert_xtralarge_ll60k.pt}}, which are widely used in speech translation. As shown in Tabel~\ref{speech_model}, as the strength of speech pre-trained models increases, the performance of the models on ST and FT improves. Intuitively, larger speech pre-trained models provide more complementary knowledge to complete the insufficient text representations.

\begin{figure}[t]
\centering
\includegraphics[width=\linewidth]{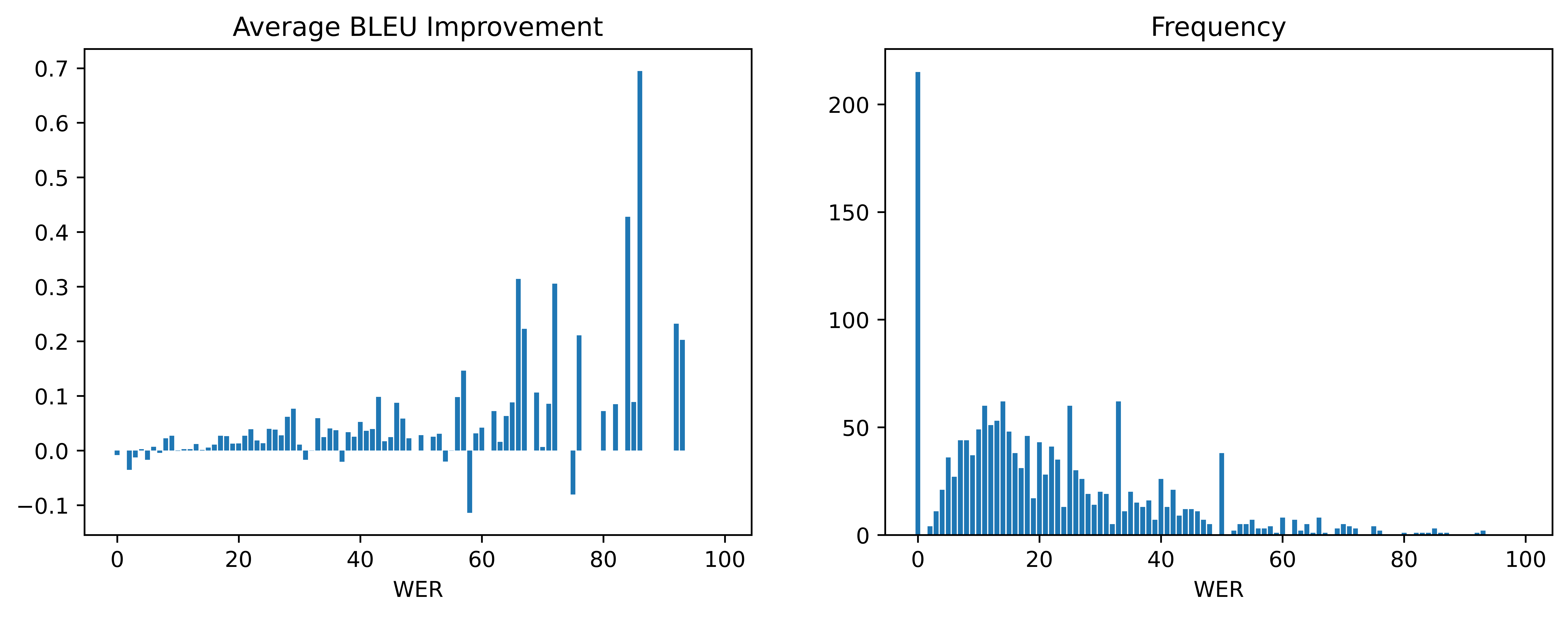}
\caption{\textbf{Left}: The average BLEU improvement for ASR transcript resulting from using fuse-speech-text on En$\rightarrow$Zh aiconf testset, under various conditions of WER. \textbf{Right}: The frequency distribution of WER on En$\rightarrow$Zh aiconf testset.}
\label{wer_improvement}
\end{figure}

\subsection{Improvement of Different WER}
We conduct an empirical study to examine the efficacy of our approach in enhancing BLEU score through the fusion of speech information under different WER present in the ASR transcript. As shown in Figure~\ref{wer_improvement}, when the word error rate of ASR transcript is minimal ($ 0 \le \text{WER} < 5$), the fusion of speech information results in a slight decrease in BLEU score. This behavior is because the transcript is adequate for achieving high translation quality, and the fusion of speech information may introduce noise. As the WER increases, the advantages of integrating speech information become more pronounced.

\subsection{Case Study}

\begin{table}[]
\centering
\resizebox{\linewidth}{!}{ 
\begin{tabular}{l|ll}
\hline
Models &  &  \\ \hline
\multicolumn{3}{c}{CASE 1} \\ \hline
\multirow{3}{*}{Ref.} & \textbf{src}: & I'm extraordinarily delight to be here. \\
~ & \textbf{asr}: & I'm extraordinary. I'd like to be here. \\
~ & \textbf{tgt}: & \begin{CJK}{UTF8}{gkai} 我非常高兴来到这里。 \end{CJK} \\ \hline
FST-MT & \textbf{tgt} & \begin{CJK}{UTF8}{gkai} 我非{\color{red} \underline{同寻常,我想在}}这里。 \end{CJK} \\
FST-ST & \textbf{tgt} & \begin{CJK}{UTF8}{gkai} {\color{blue} \deleted{我}}非常高兴来到这里。\end{CJK} \\ 
FST-FT & \textbf{tgt} & \begin{CJK}{UTF8}{gkai} 我非常高兴来到这里。\end{CJK} \\ \hline
\multicolumn{3}{c}{CASE 2} \\ \hline
\multirow{3}{*}{Ref.} & \textbf{src}: & So you could think of the viruses like, they are people. \\
~ & \textbf{asr}: & So you could think of the viruses like their people. \\
~ & \textbf{tgt}: & \begin{CJK}{UTF8}{gkai} 所以你可以把病毒想象成人。 \end{CJK} \\ \hline
FST-MT & \textbf{tgt} & \begin{CJK}{UTF8}{gkai} 所以你可以{\color{red} \underline{像他们的人民一样看待病毒。}} \end{CJK} \\
FST-ST & \textbf{tgt} & \begin{CJK}{UTF8}{gkai} 所以你可以把病毒想象成{\color{red} \underline{他们的人}}。\end{CJK} \\ 
FST-FT & \textbf{tgt} & \begin{CJK}{UTF8}{gkai} 所以你可以把病毒想象成人。\end{CJK} \\ \hline

\end{tabular}
}
\caption{Case study: cases are generated from FST-MT, FST-ST, and FST-FT on En$\rightarrow$Zh aiconf testset. The {\color{red} \underline{red underlined text}} indicates inaccurate translations, and the {\color{blue} \deleted{blue strikethrough}} indicates missing translation.}
\label{CaseStudy}
\end{table}

In this section, we present several cases generated by FST-MT, FST-ST, and FST-FT. In the first case, the ASR model mistakenly transcribes "delight" as "I'd like" due to the highly similar pronunciations of these two words. FST-MT fails to generate the correct translation as a result of errors present in the ASR transcript. Meanwhile, FST-ST produces omissions, as modeling direct speech to target translation proves to be more challenging. Notably, only FST-FT translates the sentence correctly, leveraging the complementary strengths of speech and text. In the second case, the speech signal contains numerous background noises; the ASR model mistakenly transcribes the "they are" as "their", FST-MT and FST-ST are mistranslated, and only FST-FT translates correctly. We believe this improvement comes from fusing speech and text.

\section{Conclusion}
In this paper, we propose FST, a cross-modal model which supports three distinct input modalities for translation: speech, text, and fused speech-text. We comprehensively analyze the modality gap between speech and text and utilize multiple techniques to bridge the modality gap. We then fuse speech and text to improve speech translation. Experiments and analysis demonstrate the effectiveness of our proposed method.

\section*{Limitations}
This work improves speech translation by fusing speech and text, but the model is far from being achieved for industrialgrade implementations. Although the ChatGPT and Whisper models exhibit superior speech-to-text capabilities compared to our model, we maintain that fusing speech and text remains a viable approach in the era of large-scale models. There are two significant limitations in this study that could be addressed in future research. First, our model still relies on an ASR system to transcribe speech into text, which does not address the issue of high latency in the cascade system. Second, our model needs labeled data for training, especially the <speech, transcript, target> pair. Speech data is exceptionally scarce, and obtaining speech data for many languages around the world is particularly challenging.


\bibliography{anthology}
\bibliographystyle{acl_natbib}

\appendix



\end{document}